# Deriving discriminative classifiers from generative models

Elie Azeraf, Emmanuel Monfrini, and Wojciech Pieczynski

**Abstract**—We deal with Bayesian generative and discriminative classifiers. Given a model distribution $p(x, y)$, with the observation $y$ and the target $x$, one computes generative classifiers by firstly considering $p(x, y)$ and then using the Bayes rule to calculate $p(x|y)$. A discriminative model is directly given by $p(x|y)$, which is used to compute discriminative classifiers. However, recent works showed that the Bayesian Maximum Posterior classifier defined from the Naive Bayes (NB) or Hidden Markov Chain (HMC), both generative models, can also match the discriminative classifier definition. Thus, there are situations in which dividing classifiers into "generative" and "discriminative" is somewhat misleading. Indeed, such a distinction is rather related to the way of computing classifiers, not to the classifiers themselves. We present a general theoretical result specifying how a generative classifier induced from a generative model can also be computed in a discriminative way from the same model. Examples of NB and HMC are found again as particular cases, and we apply the general result to two original extensions of NB, and two extensions of HMC, one of which being original. Finally, we shortly illustrate the interest of the new discriminative way of computing classifiers in the Natural Language Processing (NLP) framework.

## 1 Introduction

PROBABILISTIC models and related Bayesian classifiers are usually categorized into two classes: generative and discriminative. On the one hand, a generative model is usually defined by the joint probability distribution $p(x, y)$ of the hidden target variable $x$ and the observed one $y$. We can cite the Naive Bayes [1], [2], [3], the Hidden Markov Chain (HMC) [4], [5], [6], or the Gaussian Mixture Model (GMM) [7], [8]. On the other hand, a discriminative model is defined by the conditional probability $p(x|y)$ of the target $x$ given the observation $y$. We can cite the Logistic Regression [9], [10], [11], the Maximum Entropy Markov Model (MEMM) [12], or the Conditional Random Fields [13], [14].

About the classifiers, they are categorized depending on what kind of models they are based on [15]. Therefore, a probabilistic generative classifier is defined from a generative model in which one computes the joint distribution $p(x, y)$ from $p(x)$ and $p(y|x)$, and then uses the Bayes rule to compute the posterior one $p(x|y)$. Thus, the construction of a generative classifier needs $p(x)$ and $p(y|x)$. A probabilistic discriminative classifier, usually defined from a discriminative model, is based on the posterior distribution $p(x|y)$, which is given directly. Therefore, a discriminative classifier's computation neither uses $p(x, y)$ nor $p(y)$.

These definitions are currently used [14], [16], [17], [18], [19], [20], [21], [22], [23], [24], [25] and these classifier categories are largely compared with each other [14], [15], [23], [26], [27], with a general preference for the discriminative ones. One reason, especially of importance In Natural Language Processing (NLP) is that generative classifiers cannot consider arbitrary observations' features. In addition, they can be criticized for their learning strategy, often imposing to maximize the joint likelihood of a training set. These defaults are due to the joint law $p(x, y)$ computation, imposing to care about the observation's law. On their side, discriminative classifiers can consider observations' features without limitations and are generally trained by minimizing an appropriate loss function. These properties lead many authors to prefer discriminating classifiers to generative ones for classification tasks, which has led to neglect the latter in favor of the former.

This is especially the case in NLP [12], [13], [14], [28], [29], [30], [31].

The contribution of this paper is the following. After observing that all Bayesian classifiers only depend on $p(x|y)$ and are independent of the observations' distribution $p(y)$, we notice that all of them are "discriminative". However, some of them are constructed from discriminative models, while some others are constructed from generative models. Thus, the fact that the latter are called "generative" is somewhat misleading. Our contribution, provided in a general setting, consists of showing that Bayesian classifiers constructed in a generative way (using a generative model) can also be constructed in a discriminative way (using the same generative model, but using neither $p(y|x)$ nor $p(y)$). In other words, we show that many classic generative models

- *E. Azeraf is with Watson Department, IBM GBS France, 17 avenue de l'Europe, 92275 Bois-Colombes, France, and SAMOVAR, Telecom Sud-Paris, Institut Polytechnique de Paris, 9 rue Charles Fourrier, 91011 Evry, France. Email: elie.azeraf@ibm.com.*
- *E. Monfrini and W. Pieczynski are with SAMOVAR, Telecom SudParis, Institut Polytechnique de Paris, 9 rue Charles Fourrier, 91011 Evry, France. Email: {emmanuel.monfrini, wojciech.pieczynski}@telecom-sud-paris.eu.*

can produce "discriminative" classifiers, and we provide a general way of converting generative contructions of classifiers to dicriminative ones. This implies that the abandonment of certain models was not justified. For example, HMCs, known as generative models, were deemed uninteresting in NLP [12], [23]. As described in [32], Bayesian "Maximum Posterior Mode" (MPM), considered as a generative classifier, can also be computed in a discriminative way, using the original Entropic Forward-Backward algorithm instead of the classic Forward-Backward one. Consequently, HMCs are finally as interesting as discriminative models introduced to replace them [12], [13], [14]. Another example is the Naive Bayes model, discussed in [33].

This paper presents a general result, including HMC and Naive Bayes cases, specifying how to construct discriminative classifiers from generative models.

To illustrate this, let us briefly recall the Naive Bayes case, which will be developed in section 3. With this model, one considers a hidden variable $x \in \Lambda = \{\lambda_1, \ldots, \lambda_N\}$ and observations $y_{1:T} = (y_1, \ldots, y_T)$, with each $y_t$ in discrete or continuous $\Omega$. The distribution of $(x, y_{1:T})$ is given with:

$$p(x, y_{1:T}) = p(x) \prod_{t=1}^{T} p(y_t|x) \quad (1)$$

As a generative model, it allows to define a "generative" classifier:

$$\phi(y_{1:T}) = \underset{x \in \Lambda}{\operatorname{argsup}}[p(x) \prod_{t=1}^{T} p(y_t|x)] \quad (2)$$

However, as shown in [33], this classifier can also be written as a "discriminative" one:

$$\phi(y_{1:T}) = \underset{x \in \Lambda}{\operatorname{argsup}}[p(x)^{1-T} \prod_{t=1}^{T} p(x|y_t)] \quad (3)$$

Indeed, the definition of $\phi$ with (3) does use neither the joint law $p(x, y)$ nor the observations' one $p(y)$, so it matches the discriminative classifier's definition.

The organization of the paper is as follows. In the next section, we state and prove the main result, specifying how it is possible, from $p(x, y)$, to compute the posterior probability $p(x|y)$ directly, without using the laws $p(y|x)$ or $p(y)$. This result allows computing in a "discriminative" manner the classifiers considered as "generative" until now. The third section illustrates this property with some Naive Bayes based models: the classic one and two new extensions we propose, called "Pooled Markov Chain" and "Pooled Markov Chain of order 2". The fourth section deals with HMC and two of its extensions. We provide some examples of applications in NLP in section five, while conclusion and perspectives lie in the last section six.

## 2 Computing the Posterior Distribution without Using that of Observations

Let us consider a probability distribution $p(x, y)$ with a hidden variable$c = (x_1, \ldots, x_T)$ and an observed one $y = (y_1, \ldots, y_{T'})$. As we deal with classification tasks, the realization of each component $x_t$ of the hidden variable is in a discrete finite space $\Lambda$. Observation components $y_t$ are in $\Omega$, which can be discrete or continuous.

Let us recall that, according to its definition, Bayesian classifier $\phi_L: \Omega^{T'} \to \Lambda^T$ related to a loss function $L$ minimizes, among all possibles classifiers $\phi$, the mean loss $E[L(\phi(y), x]$. This is written (we note with the same lowercase letters $x$, $y$ the random variables and their realizations):

$$E[L(\phi_L(y), x] = \inf_{\phi}[E[L(\phi(y), x]] \quad (4)$$

Then one can see that $\phi_L$ verifying (4) is defined with

$$\phi_L(y) = \underset{\phi(y) \in \Lambda^T}{\operatorname{argsup}}[E[L(\phi(y), x] \,|y] \quad (5)$$

As conditional expectation in (5) only depends on conditional distribution $p(x|y)$, (5) implies that $\phi_L$ only depends on $p(x|y)$, and thus is independent from $p(y)$. Let us clarify the following point, of importance for the paper. In the literature, one distinguishes "generative" models and "discriminative" ones. In this paper, we will use the following definition:

**Definition 2.1.** *The model $p(x, y)$ is generative if the definition of $p(x, y)$ uses some $p(y_A|x_B)$, with $A, B$ nonempty subsets of {1, …, T} and {1, …, T'}, respectively. If it is not generative, it will be called "discriminative".*

**Remark 2.1**. *According to Definition 2.1, defining a discriminative model only uses distributions of the form $p(y_A), p(x_B)$ and $p(x_C|y_D)$, with $A, B, C, D$ some nonempty subsets of $\{1, \ldots, T\}$ and $\{1, \ldots, T'\}$. Therefore, a "discriminative" model within the meaning of the usual definition stating that it is defined only with the posterior law $p(x|y)$, is also "discriminative" within the meaning of Definition 2.1.*

Similarly, in the literature, Bayesian classifiers are called "discriminative" when they are defined from discriminative models, and they are called "generative" when they are defined from generative models. As noticed above, this distinction is somewhat improper as Bayesian classifiers only depend on $p(x|y)$, and thus they are all discriminative. However, they can be constructed in a generative or a discriminative way, so that discriminative or generative nature of classifiers is related to the way classifiers are contructed, not to the classifiers themselves. More precisely, let us consider:

**Definition 2.2.** *Let $\phi: y_{1:T'} \to \hat{x}_{1:T} = \phi(y_{1:T'})$ be a Bayesian classifier from a model $p(x_{1:T}, y_{1:T'})$. $\phi$ will be said "generatively constructed" if its construction calls on some $p(y_A|x_B, y_C)$, with $B$ subset of {1, …, T} and $A, C$ sub-* *sets of {1, …, T′}. It will be called "discriminatively constructed" if not.*

To clarify things, in the following we will call "generatively constructed" the classified called "generative" in the litterature. Similarly, the classified called "discriminative" in the litterature will be called "discriminatively constructed" in the paper.

This paper's main contribution is a general result allowing one to "convert" generatively constructed classifiers into discriminatively constructed ones. Then, we provide some concrete examples highlighting the interest of the conversion method proposed, especially for NLP tasks.

We denote $x_H$ the vector of hidden components of interest we want to estimate, and $\Lambda_+$ the space of all possible values of $x_H$. We set $x_O$ the vector of remaining components, and $\Lambda_-$ the space of all possible values of $x_O$.

Moreover, let $A^t$ be the minimal set of components of $y_{1:t-1}$ and of components of $x_{1:T}$ such that $p(y_t|x_{1:T}, y_{1:t-1}) = p(y_t|A^t)$. We also set $A_y^t = \{y_{t'}$ such as $y_{t'} \in A^t\}$, and $A_x^t = \{x_{t'}$ such as $x_{t'} \in A^t\}$. Therefore, $A^t = A_y^t \cup A_x^t$, and $A_y^t \cap A_x^t = \emptyset$. For example, in the Naive Bayes, we have $A^t = \{x\}$ because of $p(y_t|x, y_{1:t-1}) = p(y_t|x)$. Thus, in the Naive Bayes, we have $A_x^t = \{x\}$ and $A_y^t = \emptyset$.

In the following we deal with Bayesian classifiers corresponding to the loss function $L(x_H, x_H') = 1_{[x_H \neq x_H']}$. To simplify, $\phi_L$ will be noted $\phi$, and (5) becomes (6) below.

We can state the following:

**Proposition 2.1.** *Let $p(x, y)$ a generative model. The Bayesian generatively constructed classifier*

$$\phi(y_{1:T'}) = \underset{x_H \in \Lambda_+}{argsup}\, p(x_H|y_{1:T'}) \tag{6}$$

*is also given with (recall that for $B = \emptyset$, one has $p(A|B) = p(A)$):*

$$\phi(y_{1:T'}) = \underset{x_H \in \Lambda_+}{argsup}\Big[\sum_{x_O \in \Lambda_-} p(x_O, x_H) \prod_{t=1}^{T'} \frac{p(A_x^t|y_t, A_y^t)}{p(A_x^t|A_y^t)}\Big] \tag{7}$$

*Thus, it is also a discriminatively constructed classifier once (7) is computable with calling neither on $p(x, y)$ nor on $p(y|x)$.*

**Proof.** Let

$$\kappa(y) = \left(\prod_{t=1}^{T'} p(y_t|A_y^t)\right)^{-1}. \tag{8}$$

We have:

$$p(y|x_O, x_H) = \prod_{t=1}^{T'} p(y_t|x_O, x_H, y_{1:t-1})$$

$$= \prod_{t=1}^{T'} p(y_t|A^t) = \prod_{t=1}^{T'} \frac{p(y_t, A^t)}{p(A^t)} = \prod_{t=1}^{T'} \frac{p(y_t, A_y^t, A_x^t)}{p(A_y^t, A_x^t)}$$

$$= \prod_{t=1}^{T'} \frac{p(y_t|A_y^t)p(A_x^t|y_t, A_y^t)}{p(A_x^t|A_y^t)} = \frac{1}{\kappa(y)} \prod_{t=1}^{T} \frac{p(A_x^t|y_t, A_y^t)}{p(A_x^t|A_y^t)}.$$

Thus,

$$\kappa(y)p(x_O, x_H, y) = p(x_O, x_H) \prod_{t=1}^{T} \frac{p(A_x^t|y_t, A_y^t)}{p(A_x^t|A_y^t)},$$

which implies

$$\kappa(y)p(x_H, y) = \sum_{x_O \in \Lambda_-} p(x_O, x_H) \prod_{t=1}^{T} \frac{p(A_x^t|y_t, A_y^t)}{p(A_x^t|A_y^t)}.$$

For given $y$, maximizing $p(x_H|y_{1:T})$ is equivalent to maximize $\kappa(y)p(x_H, y)$, thus (4) and (5) are equivalent, which ends the proof. □

**Remark 2.2.** *Let us notice that many usual models, like HMCs and Naive Bayes, are written as $p(x_{1:T}, y_{1:T}) = p(x_{1:T}) \prod_{t=1}^{T'} p(y_t|x_t)$. Then $A_x^t = \{x_t\}$, $A_y^t = \emptyset$, $\kappa(y) = [\prod_{t=1}^{T} p(y_t)]^{-1}$, and $\frac{p(A_x^t|y_t, A_y^t)}{p(A_x^t|A_y^t)} = \frac{p(x_t|y_t)}{p(x_t)}$. However, as we will see below, the general form (7) is needed in more complex models like Pooled Markov Chains considered in the next section.*

**Example 2.1.** *Let us consider the HMC model, $p(x_{1:T}, y_{1:T}) = p(x_1)\, p(y_1|x_1) \prod_{t=2}^{T} p(x_t|x_{t-1})p(y_t|x_t)$. As recalled above, $A_x^t = \{x_t\}$ and $A_y^t = \emptyset$. Taking $H = (1, \dots, T)$, the classic generatively constructed classifier computing the Maxium a Posteriori is given with the well-known Viterbi algorithm [34]. According to (5), it is also written as discriminately constructed one:*

$$\phi(y_{1:T}) = \underset{x_H \in \Lambda_+}{argsup}\Big[p(x_{1:T}) \prod_{t=1}^{T} \frac{p(x_t|y_t)}{p(x_t)}\Big],$$

*and it can be computed with a method quite similar to Viterbi's one.*

## 3 Discriminative classifiers derived from Naive Bayes based models

Naive Bayes is among the most popular generative models. Given observations $y_{1:T}$, with $y_t \in \Omega$, the distribution $p(x, y_{1:T})$ of the Naive Bayes is given with (1), and the generatively constructed classifier is given with (2).

Naive Bayes can be used for many tasks as text classification [35] or sentiment analysis [23]. However, using its usual classifier (2) for these tasks is not relevant. Indeed, it can neither efficiently consider arbitrary observations' features [15], [23], such the handcrafted ones (suffixes, prefixes, …), nor the numerical vector returns by an embedding method such as BERT [36], Flair [37], or XLNet [38].

This section contains the following contributions. We extend, in a generative manner, Naive Bayes to two new models called Pooled Markov Chains (Pooled MCs) and Pooled Markov Chains of order 2 (Pooled MC2s). Then, we apply Proposition 2.1 to show (3) for Naive Bayes and the discriminative forms of Bayesian classifiers based on Pooled MCs and Pooled MC2s.

We give in Figure 1 and Table 1 the probabilistic oriented graphs and the joint law $p(x, y_{1:T})$ of the three models. Let us notice that, in the Pooled MC, the distribution $p(y_{1:T}|x)$ is Markov distribution, while in Pooled MC2, it is a second order Markov distribution.

In the three models, the Bayesian generatively constructed classifier is defined with

$$\phi(y_{1:T}) = \underset{x \in \Lambda}{\text{argsup}}[p(x)p(y_{1:T}|x)].$$

To apply Proposition 2.1 for defining the three classifiers in a discriminative form, we must compute $A^t$, $A_x^t$, and $A_y^t$. Easily determined from the dependence graphs in Figure 1, they are specified in Table 2.

Finally, applying (5) to Naive Bayes gives (3), applying (7) to Pooled MC gives

$$\phi(y_{1:T}) = \underset{x \in \Lambda}{\text{argsup}}[p(x|y_1)\prod_{t=1}^{T-1}\frac{p(x|y_t, y_{t-1})}{p(x|y_{t-1})}], \quad (10)$$

and applying (7) to second order Pooled MC gives

$$\phi(y_{1:T}) = \underset{x \in \Lambda}{\text{argsup}}[p(x|y_2, y_1)\prod_{t=1}^{T-2}\frac{p(x|y_t, y_{t-1}, y_{t-2})}{p(x|y_{t-1}, y_{t-2})}]. \quad (11)$$

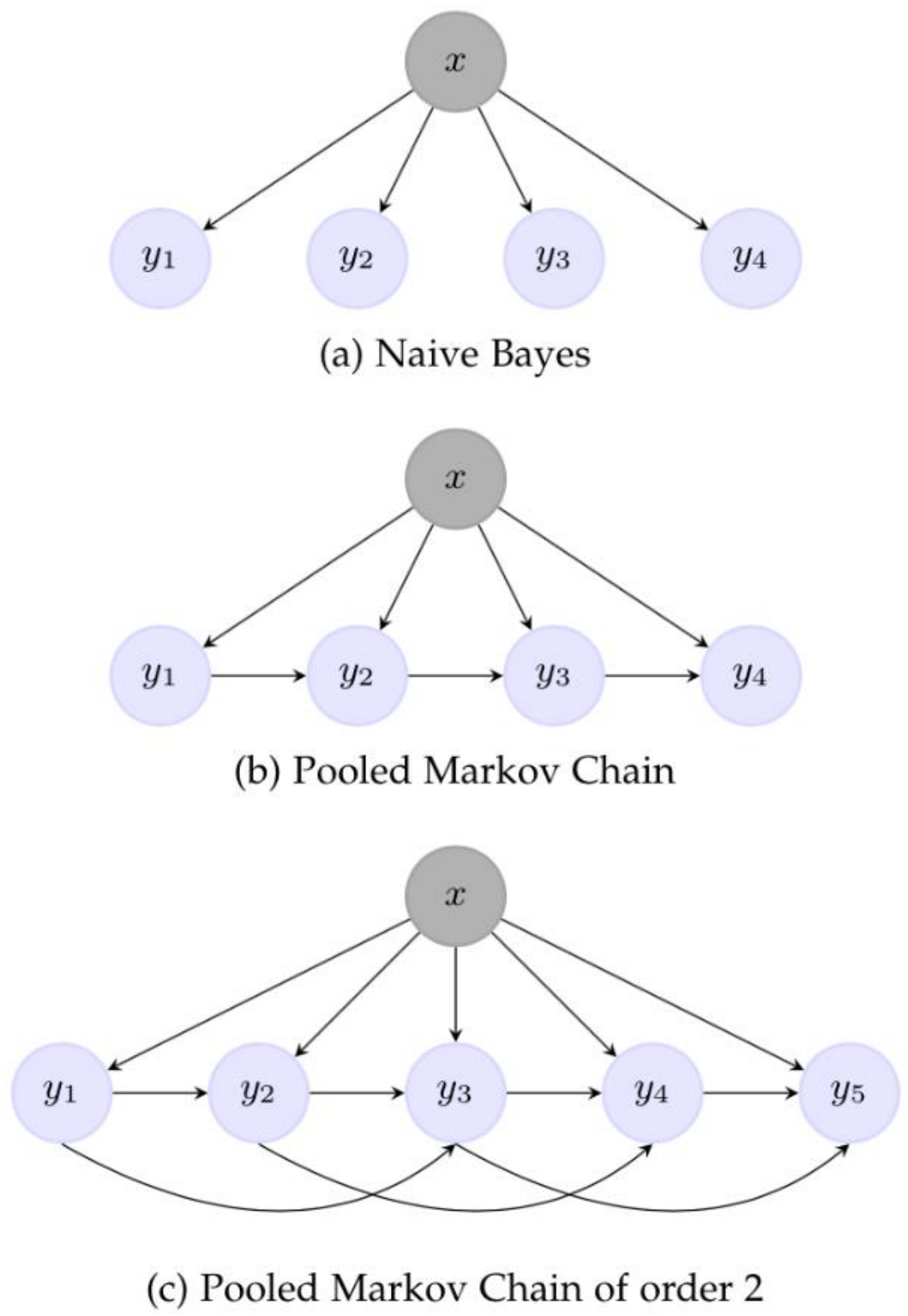

Fig. 1: Probabilistic graphs of the Naive Bayes based models

| Model | Law |
|---|---|
| Naive Bayes | $p(x)\prod_{t=1}^{T} p(y_t|x)$ |
| Pooled MC | $p(x)p(y_1|x)\prod_{t=1}^{T-1} p(y_{t+1}|x, y_t)$ |
| Pooled MC2 | $p(x)p(y_1|x)p(y_2|x, y_1)\prod_{t=1}^{T-2} p(y_{t+2}|x, y_t, y_{t+1})$ |

TABLE 1: Law of $p(x, y_{1:T})$ for Naive Bayes based models

| | $A^t$ | $A_x^t$ | $A_y^t$ |
|---|---|---|---|
| Naive Bayes | $\{x\}$ | $\{x\}$ | $\emptyset$ |
| Pooled MC | $\{x, y_{t-1}\}$ | $\{x\}$ | $\{y_{t-1}\}$ |
| Pooled MC2 | $\{x, y_{t-1}, y_{t-2}\}$ | $\{x\}$ | $\{y_{t-1}, y_{t-2}\}$ |

TABLE 2: Notations of Section 2.1 for Naive Bayes based models

# 4 Discriminative classifiers derived from Hidden Markov Chain s based models

The Hidden Markov Chain is another popular generative model. A couple of stochastic processes forms it: the observed $y_{1:T}$ and the hidden $x_{1:T}$, with for each $t \in \{1, \dots, T\}, y_t \in \Omega, x_t \in \Lambda = \{\lambda_1, \dots, \lambda_N\}$. Its distribution is given in Table 3.

This sequential model [4], [5], [6] admits huge areas of applications; let us mention image segmentation [39] and automatic speech recognition [40] as examples. Here we focus on some of its applications in NLP. Authors used it in different tasks as Part-Of-Speech tagging [41], Chunking [42], or Named Entity Recognition [43]. However, computed in the generative way, related Bayesian classifiers were unable to consider arbitrary features, so HMC was neglected for twenty years for NLP applications. According to the results presented, abandoning HMCs in NLP applications was not necessary. On the contrary, they can be extended in different directions keeping the same abilities as discriminative models designed to replace them. We propose two extensions of the HMC: the HMC or order 2 (HMC2) and the HMC with a direct correlation between a hidden variable and the following observed one (HMC+), and we show their applicability in NLP. The dependences graphs and distribution of HMC based models are presented in Figure 2 and Table 3, respectively.

Let us consider the Bayesian Maximum Posterior Mode (MPM) classifier, which consists of maximizing $p(x_t|y_{1:T})$ for each $t \in \{1, \dots, T\}$. In other words, we consider $H = \{t\}$. Usually, MPM is computed in a generatively constructed using the Forward-Backward algorithm [5], [40], [42]. As mentioned above, a discriminative way of computing MPM with the Entropic Forward-Backward (EFB) algorithm has been recently proposed in [32]. Here, by applying Proposition 2.1, we find EFB again in the HMC case. Then we apply Proposition 2.1 to compute original discriminatively constructed MPMs in HMC2 and HMC+ cases.

Let us specify how discriminatvely constructed

MPM are obtained in the three cases considered. According to the MPM principle, $\phi(y_{1:T}) = (\widehat{x_1}, \ldots, \widehat{x_T}) = (\phi_1(y_{1:T}), \ldots, \phi_T(y_{1:T}))$. Thus, we apply (7) $T$ times, with $H = \{1\}$, …, $H = \{T\}$. Therefore, $\Lambda_+ = \Lambda$ and $\Lambda_- = \Lambda^{T-1}$. Moreover, $A_y^t = \emptyset$ in the three models HMC, HMC2, and HMC+, and $A_x^t$ equals $\{x_t\}, \{x_t\}$, and $\{x_{t-1}, x_t\}$, for HMC, HMC2, and HMC+, respectively.

Let us consider the HMC case. According to (5), each $\phi_h(y_{1:T})$ is written

$$\phi_h(y_{1:T})$$
$$= \operatorname*{argsup}_{x_h \in \Lambda} [\sum_{x_O \in \Lambda_-} p(x_1)p(x_2|x_1) \ldots p(x_T|x_{T-1}) \prod_{t=1}^{T} \frac{p(x_t|y_t)}{p(x_t)}]$$
$$= \operatorname*{argsup}_{x_h \in \Lambda} [\sum_{x_1} p(x_1) \frac{p(x_1|y_1)}{p(x_1)} \sum_{\substack{x_{t+1} \\ t+1 \neq h}} p(x_{t+1}|x_t) \frac{p(x_{t+1}|y_{t+1})}{p(x_{t+1})}]$$
$$= \operatorname*{argsup}_{x_h \in \Lambda} [\sum_{x_1} p(x_1|y_1) \sum_{\substack{x_{t+1} \\ t+1 \neq h}} p(x_{t+1}|x_t) \frac{p(x_{t+1}|y_{t+1})}{p(x_{t+1})}]. \quad (10)$$

Thus, we retrieve the Entropic Forward-Backward algorithm proposed in [32]. Indeed, (10) verifies

$$\phi_t(y_{1:T}) = \operatorname*{argsup}_{x_h \in \Lambda} [\alpha_h(x_h)\beta_h(x_h)],$$

where Entropic Forward quantities $\alpha_1(x_1)$, …, $\alpha_T(x_T)$, and Entropic Backward ones $\beta_1(x_1)$, …, $\beta_T(x_T)$ are computed with the following forward and backward recursions:

$\alpha_1(x_1) = p(x_1|y_1)$;

$\alpha_{u+1}(x_{u+1}) = \frac{p(x_{u+1}|y_{u+1})}{p(x_{u+1})} \sum_{x_u \in \Lambda_X} p(x_{u+1}|x_u)\alpha_u(x_u)$ ;

$\beta_T(x_T) = 1$;

$\beta_u(x_u) = \sum_{x_{u+1} \in \Lambda} \frac{p(x_{u+1}|y_{u+1})}{p(x_{u+1})} p(x_{u+1}|x_u)\beta_{u+1}(x_{u+1})$.

Let us consider the HMC2 case. We have:

$$\phi_h(y_{1:T}) = \operatorname*{argsup}_{x_h \in \Lambda} [\sum_{x_O \in \Lambda_-} p(x_1)p(x_2|x_1)p(x_3|x_1, x_2) \ldots$$
$$\times p(x_T|x_{T-1}, x_{T-2}) \prod_{t=1}^{T} \frac{p(x_t|y_t)}{p(x_t)}]$$
$$= \operatorname*{argsup}_{x_h \in \Lambda} [\sum_{x_1} p(x_1|y_1) \sum_{x_2} p(x_2|x_1) \frac{p(x_2|y_2)}{p(x_2)}$$
$$\times \sum_{\substack{x_{t+2} \\ t+2 \neq h}} p(x_{t+2}|x_t, x_{t+1}) \frac{p(x_{t+1}|y_{t+1})}{p(x_{t+1})}] \quad (13)$$

Finally, let us consider the HMC+ case. We have:

$$\phi_t(y_{1:T})$$
$$= \operatorname*{argsup}_{x_h \in \Lambda} [\sum_{x_O \in \Lambda_-} p(x_1)p(x_2|x_1) \ldots p(x_T|x_{T-1}) \prod_{t=1}^{T} \frac{p(x_{t-1}, x_t|y_t)}{p(x_{t-1}, x_t)}]$$
$$= \operatorname*{argsup}_{x_h \in \Lambda} [\sum_{x_1} p(x_1|y_1) [\sum_{\substack{x_{t+1} \\ t+1 \neq h}} \frac{p(x_t, x_{t+1}|y_{t+1})}{p(x_t)}]] \quad (14)$$

As for the HMC, (13) and (14) can be written as a special case of the EFB algorithm for HMC2 and HMC+ cases, respectively.

| Model | Law |
|---|---|
| HMC | $p(x_1) \prod_{t=1}^{T-1} p(x_{t+1}\|x_t) \prod_{t=1}^{T} p(y_t\|x_t)$ |
| HMC2 | $p(x_1)p(x_2\|x_1) \prod_{t=1}^{T-2} p(x_{t+2}\|x_t, x_{t+1}) \prod_{t=1}^{T} p(y_t\|x_t)$ |
| HMC+ | $p(x_1) \prod_{t=1}^{T-1} p(x_{t+1}\|x_t)p(y_1\|x_1) \prod_{t=1}^{T-1} p(y_{t+1}\|x_t, x_{t+1})$ |

TABLE 3: Law of $p(x_{1:T}, y_{1:T})$ for Hidden Markov Chain based models

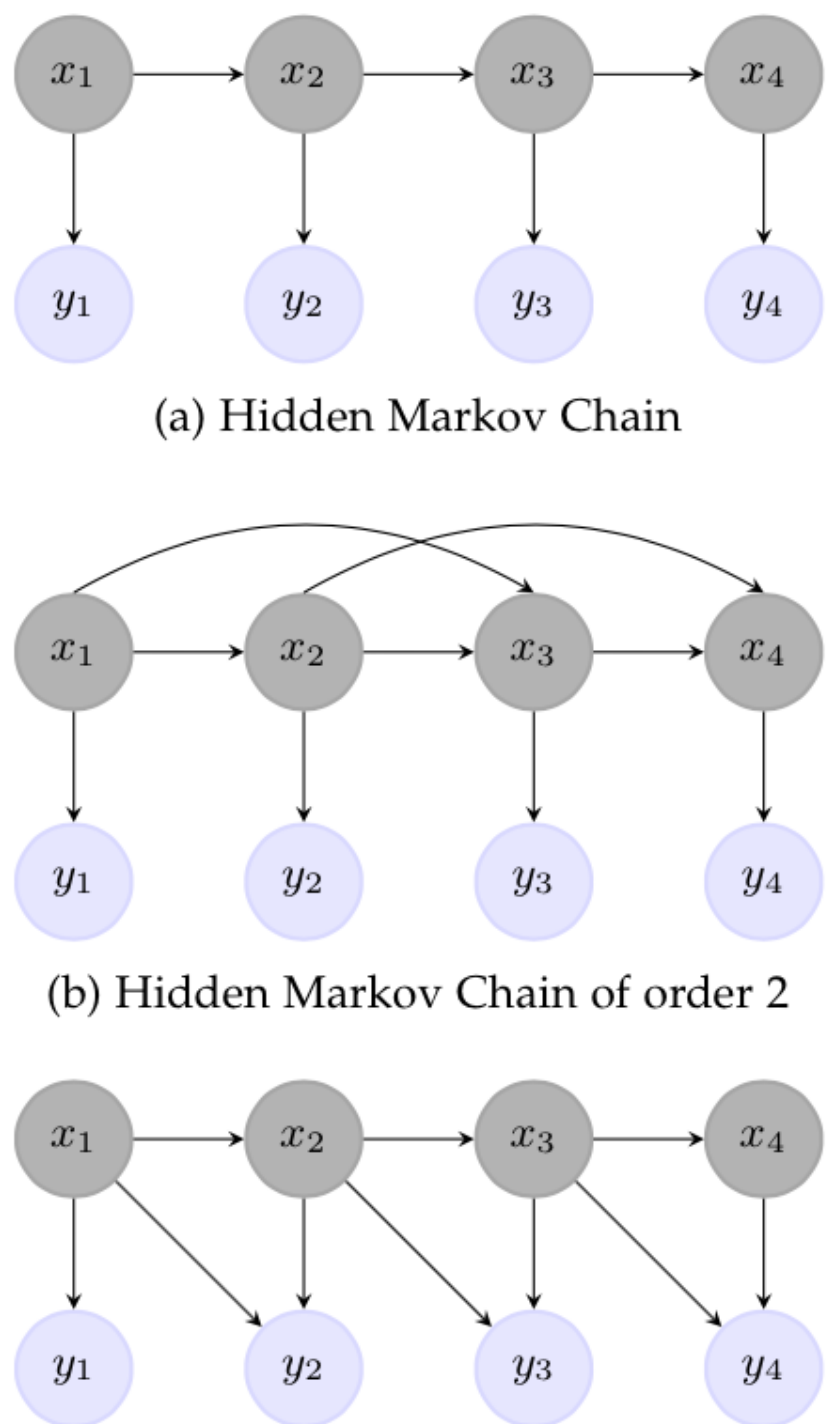

(a) Hidden Markov Chain

(b) Hidden Markov Chain of order 2

(c) Hidden Markov Chain with direct correlation between $x_t$ and $y_{t+1}$

Fig. 2: Probabilistic graphs of Hidden Markov Chain based models

## 5 Generatively and Discriminatively Constructed Classifiers in NLP

This short section is devoted to some examples illustrating advantage of discriminatively constructed classifiers over generatively contructed ones in some tasks related to NLP.

On the one hand, we apply the Naive Bayes classifiers to Text Classification on AG News dataset and to Sentiment Analysis on IMDB [46] dataset, with GloVe and FastText embeddings. On the other hand, we apply the HMC ones to Part-Of-Speech tagging on Universal Dependency English dataset [47] and Named-Entity Recognition on CoNLL 2003 dataset [48], with Flair and BERT embeddings. The code is written in python with PyTorch [49] and Flair [50] librairies, using gradient descent with the Adam [51] optimizer for parameter learning of discriminative classifiers. Results are in Tables 4 and 5. They highlight the importance of being able to consider arbitrary features without limitations. Indeed, the performances achieved by the generatively constructed classifiers are irrelevant for every task or embedding, unlike the ones achieved by the discriminatively constructed classifiers.

| | | GloVe | FastText |
|---|---|---|---|
| Text Classification | Gen | 39.78% | 39.89% |
| | Dis | $\mathbf{87.15}\% \pm 0.17$ | $\mathbf{89.10}\% \pm 0.21$ |
| Sentiment Analysis | Gen | 53.48% | 53.10% |
| | Dis | $\mathbf{74.51}\% \pm 2.71$ | $\mathbf{82.50}\% \pm 0.46$ |

TABLE 4: Generatively constructed and discriminative classifieurs of the Naive Bayes used for Text Classification and Sentiment Analysis on AG News and IMDB datasets with GloVe and FastText embeddings

| | | BERT | Flair |
|---|---|---|---|
| POS Tagging | Gen | 20.25% | 24.07% |
| | Dis | $\mathbf{93.42}\% \pm 0.07$ | $\mathbf{95.20}\% \pm 0.04$ |
| NER | Gen | 0 | 0 |
| | Dis | $\mathbf{87.32} \pm 0.04$ | $\mathbf{87.81} \pm 0.11$ |

TABLE 5: Generatively constructed and discriminative classifieurs of the Hidden Markov Chain used for Part-Of-Speech (POS) tagging and Named-Entity-Recognition (NER) on Universal Dependencies and CoNLL 2003 datasets with BERT and Flair embeddings

## 6 Conclusion and perspectives

We studied how to convert classifiers constructed in a generative way – called "generative classifiers" in the literature – into classifiers constructed in a discriminative way – called "discriminative classifiers" in the literature. We stated a general result, and we showed how it is applied to the Naive Bayes model and two of its extensions we have proposed. We also applied it to the classic Hidden Markov Chain, retrieving results presented in [32], and we extend it for two generalizations of HMs.

The presented results show that on contrary to what is generally considered, probabilistic generative models allow defining discriminative classifiers, which can handle arbitrary observation features. This point is of decisive interest, as the generatively contructed classifiers cannot consider arbitrary features without some strong conditions. It is damaging in some fields like NLP, where embedding methods convert word to numerical vector of large sizes. We can cite BERT [36], resulting on vectors of size 784 per word, Flair [37] of size 4096, FastText [44] of size 300, or even GloVe [45] of size 100.

Therefore, a promising perspective lies in applying generative probabilistic models to tasks that are in general considered unsuitable for them. This can encourage to put back generative models at the front of the scene for some tasks, like those current in NLP.

Moreover, one can model the different parameters induced in (7) with neural network functions [52, 53]. A first example, with the specific HMC's case, is presented in [54]. By generalizing this approach, it will be possible to define new neural architectures from probabilistic models. It will allow taking advantage of the probabilistic and the neural network frameworks assembled for Machine Learning applications.

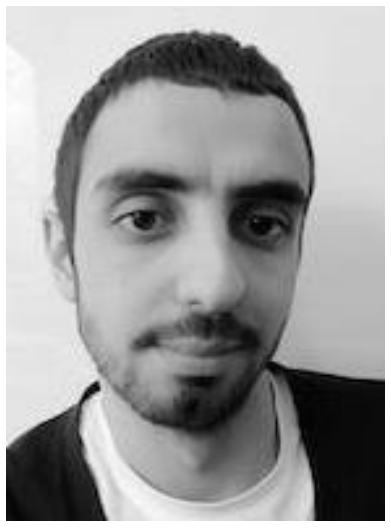

**Elie Azeraf** received M.Sc. degrees in statistics from Télécom SudParis, Evry, France, and ENSAE Paris, Palaiseau, France, in 2018. He is currently pursuing the Ph.D. degree in the CITI Department, Telecom SudParis, Institut Polytechnique de Paris. His research interests include Machine Learning, statistics, Hidden Markov Chains, probabilistic models, neural networks, and Natural Language Processing.

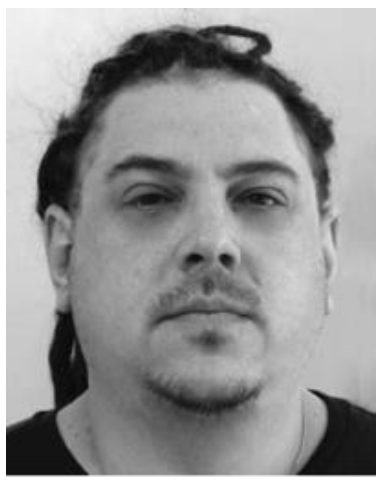

**Emmanuel Monfrini** received the Ph.D. degree in mathematical statistics from Paris VI University, Paris, France, in 2002. He has been a Professor at Télécom SudParis, Institut Polytechnique de Paris, Evry, France, since 2008. His research interest is centered around statistical modeling, with a particular interest for Markovian models and their application in several fields including signal and image processing.

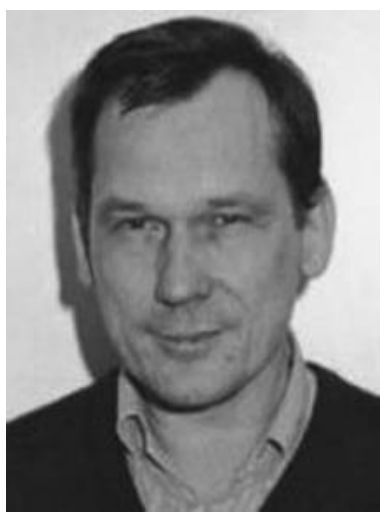

**Wojciech Pieczynski** received the Doctorat d'État degree in mathematical statistics from the Université Pierre et Marie Curie, Paris, France, in 1986. He is currently a Professor at the Telecom SudParis, Institut Polytechnique de Paris, Evry, France, and Head of the Communications, Images, and Information Processing (CITI) department. His research interests include mathematical statistics, stochastic processes, and statistical signal processing.